\documentclass{article}
\usepackage[utf8]{inputenc}
\usepackage[T1]{fontenc}
\usepackage{authblk}
\usepackage{setspace}
\usepackage[margin=1.25in]{geometry}
\usepackage{graphicx}
\usepackage{subcaption}
\usepackage{placeins}
\usepackage{amsmath,amssymb}
\usepackage{booktabs,tabularx}
\usepackage{xcolor}
\usepackage[hidelinks]{hyperref}

\usepackage[style=nejm,
citestyle=numeric-comp,
sorting=none,
giveninits=true,
maxbibnames=10,
minbibnames=10,
isbn=false,
url=false,
eprint=false,
urldate=long,
doi=true,
articledoi=true]{biblatex}
\DeclareFieldFormat{labelnumberwidth}{\mkbibbrackets{#1}}
\DeclareFieldFormat*{pages}{#1}
\renewbibmacro*{volume+number+eid}{%
  \printfield{volume}%
  \printfield[parens]{number}%
  \setunit{\addcolon}%
  \printfield{eid}}
\DeclareDelimFormat{finalnamedelim}{\addcomma\space}
\DeclareDelimFormat{andothersdelim}{\addcomma\space}
\DefineBibliographyStrings{english}{andothers={et al\adddot}}
\renewbibmacro*{journal+issuetitle}{%
  \usebibmacro{journal}%
  \setunit*{\adddot\space}%
  \usebibmacro{issue+date}%
  \setunit*{\addsemicolon}%
  \usebibmacro{volume+number+eid}%
  \newunit}
\renewbibmacro*{byeditor+others}{}
\DeclareFieldFormat{url}{\url{#1}}
\DeclareFieldFormat{doi}{\url{https://doi.org/#1}}
\DeclareFieldFormat{booktitle}{#1}
\DeclareSourcemap{
  \maps[datatype=bibtex]{
    \map{
      \step[fieldsource=archiveprefix, match=\regexp{\A[Aa][Rr][Xx][Ii][Vv]\Z}, final]
      \step[fieldset=howpublished, fieldvalue={arXiv}]
    }
  }
}
\newcommand*{\spjmonth}[1]{\ifcase#1\or Jan\or Feb\or Mar\or Apr\or May\or Jun\or Jul\or Aug\or Sep\or Oct\or Nov\or Dec\fi}
\DeclareBibliographyDriver{inproceedings}{%
  \usebibmacro{bibindex}%
  \usebibmacro{begentry}%
  \usebibmacro{author/translator+others}%
  \setunit{\printdelim{nametitledelim}}\newblock
  \usebibmacro{title}%
  \newunit\newblock
  \printtext{Paper presented at:}\setunit{\addspace}%
  \printfield{booktitle}%
  \iffieldundef{series}{}{%
    \setunit{\addspace}%
    \printtext[parens]{\printfield{series}%
      \iffieldundef{volume}{}{\addcomma\space vol\adddot\space\thefield{volume}}}}%
  \iffieldundef{series}{\iffieldundef{volume}{}{\setunit{\addcomma\space}\printtext{vol\adddot\space\thefield{volume}}}}{}%
  \setunit{\addsemicolon\space}%
  \printfield{year}%
  \newunit
  \iffieldundef{pages}{}{\printtext{p\adddot\space\printfield{pages}}}%
  \newunit\newblock
  \usebibmacro{doi+eprint+url}%
  \usebibmacro{finentry}}
\DeclareBibliographyDriver{online}{%
  \usebibmacro{bibindex}%
  \usebibmacro{begentry}%
  \usebibmacro{author/editor+others/translator+others}%
  \setunit{\printdelim{nametitledelim}}\newblock
  \usebibmacro{title}%
  \newunit\newblock
  \printfield{note}%
  \newunit\newblock
  \printlist{organization}%
  \newunit\newblock
  \iffieldundef{day}{}{\printtext{\thefield{day}\space\spjmonth{\thefield{month}}\space}}%
  \printfield{year}%
  \newunit\newblock
  \iffieldundef{urlyear}{}{\printtext{[accessed \thefield{urlday}\space\spjmonth{\thefield{urlmonth}}\space\thefield{urlyear}]}}%
  \setunit{\addspace}%
  \printfield{url}%
  \usebibmacro{finentry}}

\title{ReSkill: Selective Self-Improvement for Robot Manipulation}

\author[1,*]{Zheng SUN}
\author[1,*]{Zhihao LI}
\author[1,*]{Zhuo LI}
\author[1]{Wenrui LIU}
\author[2]{Xin WANG}
\author[1,$\dagger$]{Fei~CHEN}
\affil[1]{The Chinese University of Hong Kong, Hong Kong SAR, China}
\affil[2]{Openmind Intelligent Robot Co., Ltd., Wuhu, China}
\affil[*]{These authors contributed equally to this work.}
\affil[$\dagger$]{Corresponding author: \href{mailto:feichen@cuhk.edu.hk}{feichen@cuhk.edu.hk}}

\date{\vspace{-1.5ex}\normalsize Project page: \url{https://zainzh.github.io/ReSkill/}}
\begin{document}

\maketitle

\begin{abstract}

Self-improvement on compositional manipulation tasks requires deciding what to reuse, where to improve, and which changes to retain.
We present ReSkill, a selective self-improvement framework that concentrates revision on capability gaps.
A coding agent constructs an object-centric task graph from an instruction and a skill library, then uses node-level execution records to locate failures.
At the task level, it revises stage organization, hand-off conditions, and recovery routes; at the skill level, it strengthens perception, motion, manipulation, and verification capabilities that repeatedly fail.
Candidate revisions undergo full-task evaluation and paired comparisons before adoption, allowing local changes to accumulate into validated skills and an independently executable policy.
The agent participates only during development; the deployed policy calls the retained skills without invoking the agent at run time.
On the same development set, average success across ten RoboDojo tasks rises from 30.0\% for the initial versions to 86.2\% for the final adopted programs; retained programs achieve 83.6\% in recorded robustness and model-selection evaluations. Across eight real-world tasks, ReSkill succeeds in 72 of 80 rebuilt-simulation trials and 62 of 80 real-robot trials (77.5\%).
The median revision changes only three functions, approximately 0.5\% of a program, while 39\% of task-specific functions in later-added tasks directly reuse existing implementations.
These results show how locating local deficiencies, validating candidate changes, and accumulating reusable skills can improve execution of compositional manipulation tasks.

\end{abstract}

\section{Introduction}
\label{sec:intro}

When a robot encounters a new manipulation task, the capability it lacks often covers only part of the task.
A compositional task may involve observation, approach, grasping, transport, placement, and return; existing controllers or skills can already handle many of these stages, while a few fail on new objects, layouts, occlusions, or execution conditions.
Robot foundation models have expanded robots' ability to act on visual and language inputs \cite{brohan2023rt2,black2024pi0}, while language-model agents have enabled flexible task decomposition and skill coordination \cite{ahn2022saycan,liang2023cap}.
Nevertheless, new multi-stage tasks are often treated as unified learning or adaptation problems.
This can require relearning capabilities the robot already has and increase the costs of data, computation, and real-world interaction.
This raises a more fundamental question: when a robot encounters a new compositional manipulation task, what exactly should it improve?

Existing research addresses this problem through end-to-end policy learning, skill composition, and agent self-improvement, but still lacks explicit modeling of where capability gaps occur.
End-to-end policies map observations and task instructions directly to actions \cite{brohan2023rt2,black2024pi0}, making it difficult to distinguish stages that already work reliably from those that need improvement.
Agentic manipulation systems based on task decomposition, programmatic control, and skill calls \cite{ahn2022saycan,liang2023cap,singh2023progprompt,huang2022innermonologue} flexibly compose existing capabilities, but mainly improve how those capabilities are organized and used: when a skill repeatedly fails under new conditions, replanning or calling the same skill again does not change the skill itself.
Voyager accumulates executable skills, Reflexion stores reflections from task feedback, and Self-Refine iteratively critiques and revises model outputs \cite{wang2023voyager,shinn2023reflexion,madaan2023selfrefine}.
For robot manipulation, candidate revisions must also be assessed across layouts: an improvement on a few development scenes may not persist under different initial conditions.
The unresolved question is therefore: how can a robot first locate the capability gaps in a task, improve only those parts, and reliably judge whether the improvement is real and worth retaining and reusing?

\begin{figure}[!htbp]
\centering
\includegraphics[width=\linewidth]{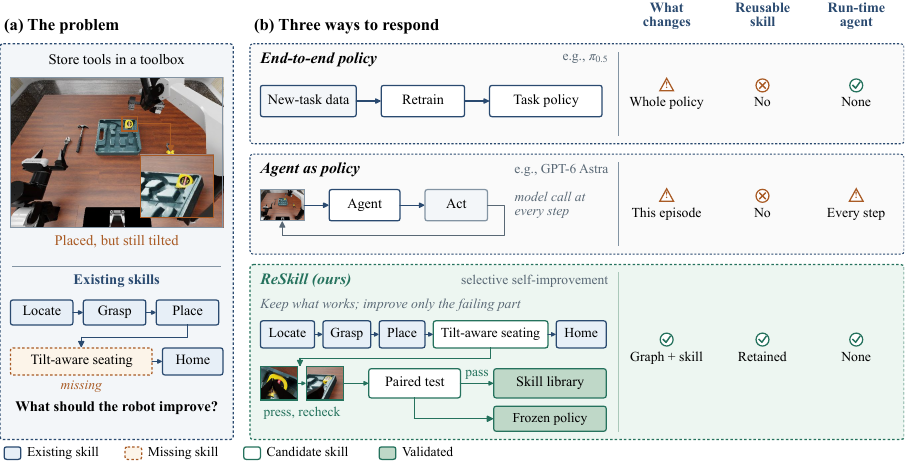}
\caption{What should a robot improve for a new manipulation task? (a) In tool storage, the existing locate, grasp, and place skills succeed, but the tape measure remains tilted because a tilt-aware seating skill is missing. (b) An end-to-end policy retrains the whole task policy, and an agent used as the policy calls a model at every step without retaining a reusable skill. ReSkill keeps the parts that work and, during development, revises the task graph and local skills, validates candidates through full-task and paired evaluation, and retains reusable skills and a frozen policy that makes no run-time agent calls. The illustrated case adds a tilt-aware seating skill.}
\label{fig:teaser}
\end{figure}

We introduce ReSkill, a framework for selective self-improvement on compositional manipulation tasks (Figure~\ref{fig:teaser}).
The robot first identifies the task structure and assesses its existing capabilities, then concentrates improvement on the gaps that remain.
A coding agent serves as the learning orchestrator, constructing an object-centric task graph from a task description and, optionally, one demonstration.
Each node specifies a skill call, its object roles, entry and completion conditions, and budget; edges specify success, failure, and recovery routes.
Within this graph, explicit geometric routines handle stages that can be programmed directly, and library skills cover interactions the robot already knows how to perform.
Repeated execution failures reveal the unresolved stages or skills and direct the agent's revisions, allowing development to preserve and build on capabilities that already work.

Improvement proceeds at two scales, using the same node-level execution records and visual traces.
At the task level, the agent locates the earliest stage that deviates and revises stage decomposition, hand-off conditions, skill bindings, or recovery procedures.
For example, in cup nesting it changes the release-stage completion condition to require both release and the intended nesting relation, while keeping the grasping and insertion skills unchanged.
At the skill level, improvement is organized around four types of reusable capability: perception skills estimate object geometry, motion skills generate feasible trajectories, manipulation skills perform interactions such as grasping and placement, and verification skills check outcomes and guide correction.
Failure evidence accumulated across tasks identifies which skill needs strengthening; validated improvements are shared through the skill library or transferred into other task programs.
New tasks therefore start from the existing task with the closest skill requirements.

Every revision at either scale must pass multi-stage evaluation before adoption.
Candidates are screened on a few development layouts, evaluated on the full development set and robustness layouts, and compared with the incumbent on paired test layouts available only to the orchestrator.
When results are close, paired evaluations are repeated on layouts where outcomes disagree.
This links local revision to evidence that determines whether an improvement should be retained and reused.

We evaluate ReSkill on ten RoboDojo simulation tasks \cite{chen2026robodojo} and eight real-world tasks on the Xtrainer platform.
For each real-world task, a single teleoperated demonstration is used to rebuild the corresponding scene in MuJoCo (real-to-sim); development takes place in the rebuilt environment, and the resulting program is deployed on the real robot (sim-to-real).
This process supplies a development environment for real-world tasks and is not itself part of the method.
The coding agent participates only during development, revising task graphs and skills, submitting evaluations, and deciding which revisions to adopt; the deployed program is frozen and makes no external large-model calls at run time.
Model selection uses test layouts accessed by the orchestrator.
On RoboDojo, retained programs make zero run-time API calls and achieve an average success rate of 83.6\% across the ten tasks in 337 recorded evaluation episodes.
During development, average task success rises from 30.0\% for the initial task graphs to 86.2\%.
On the eight real-world tasks, ReSkill succeeds in 72 of 80 trials in rebuilt simulation and 62 of 80 trials on the real robot.
We further analyze revision locality, the evolution of failure types, cross-task skill reuse and how quickly new tasks get started, the role of a single demonstration, and development-agent model usage.

Our main contributions are as follows.

\paragraph{Selective Self-Improvement via Task Graphs}
We introduce a selective self-improvement framework that organizes stages, skills, and recovery routes in an object-centric task graph and focuses improvement on the stages and skills that actually fail during execution, rather than relearning the entire task.

\paragraph{Two-Scale Improvement: Task-Graph Revision and Skill Reuse}
We unify failure-driven task-graph revision with skill organization, improvement, and reuse in a single development loop, and provide a multi-stage evaluation and adoption protocol centered on paired comparisons on model-selection test layouts to assess which revisions improve task execution.

\paragraph{Empirical Study of How Skills Improve, Transfer and Pay Off}
We evaluate the framework on ten RoboDojo simulation tasks and eight real-world tasks on the Xtrainer platform, and use more than ten thousand verified simulation episodes and code snapshots from 774 revisions to quantify revision locality, the evolution of failure types, function-level skill reuse, the role of demonstrations, and development cost.

\FloatBarrier
\subsection{Related work}
\label{sec:related}

\subsubsection{Agentic and self-improving robot systems}

Recent robot systems increasingly use high-level agents to decompose complex tasks into combinations of programmatic operations and learned skills.
SayCan and Code as Policies demonstrate language models' ability to decompose tasks and invoke skills \cite{ahn2022saycan,liang2023cap}; the more recent Harness VLA \cite{zhang2026harnessvla} combines a frozen vision--language--action model with analytic primitives and extends the policy's effective range through re-grounding, staging, and skill orchestration.
ASPIRE, Zetta, and RISE explore robot self-improvement through programmatic skill accumulation, failure recovery, and policy-level optimization, respectively \cite{lu2026aspire,ding2026zetta,yang2026rise}.
ReSkill couples task-graph revision with local skill improvement: execution records identify whether a failure calls for a change in task organization or skill implementation, and paired evaluation determines which updates are retained for later tasks.
Our central questions are where improvement is actually needed and how improvements can be reliably retained and reused.

\subsubsection{Skill libraries and code as policies}

Earlier work on long-horizon bimanual manipulation also used staged learning: Mixline learns subtasks with online reinforcement learning and combines them through offline learning from the resulting trajectories \cite{sun2022mixline}.
Programs provide a natural representation for organizing robot skills: Code as Policies and ProgPrompt use language models to write programs that call perception and control primitives \cite{liang2023cap,singh2023progprompt}; Inner Monologue returns execution feedback to the planner as text \cite{huang2022innermonologue}; Voyager stores generated skill code in a retrievable library for later tasks \cite{wang2023voyager}; and Eureka uses a coding agent to iteratively rewrite reward functions \cite{ma2024eureka}.
ReSkill likewise represents skills and task graphs as programs, but does not generate them on the fly at deployment: programs are repeatedly executed during development, failures are attributed to stages, and programs are frozen after multi-stage evaluation and remain unchanged during deployment.
We also measure at the code level how extensively skills are reused verbatim or with rewriting across tasks (Section~\ref{sec:exp-reuse}), a quantity rarely reported systematically in prior work.

\subsubsection{Agent-driven iterative improvement}

Reflexion uses task feedback to guide subsequent attempts \cite{shinn2023reflexion}, and Self-Refine iteratively critiques and revises an output using feedback generated by the same model \cite{madaan2023selfrefine}, while The AI Scientist delegates the propose--implement--experiment--summarize process to a model \cite{lu2024aiscientist}.
Robot-task feedback varies across layouts, and development layouts are limited; relying on those layouts alone can mistake fitting specific scenes for a capability improvement.
We therefore treat improvement during development as a version-selection problem, using multiple layout splits, progressively stricter adoption rules, and paired comparisons on the model-selection test layouts to decide which revisions to retain (Section~\ref{sec:method-revision}).

\FloatBarrier
\section{Methods}
\label{sec:methods}
\label{sec:method}

\begin{figure}[!htbp]
\centering
\includegraphics[width=\linewidth]{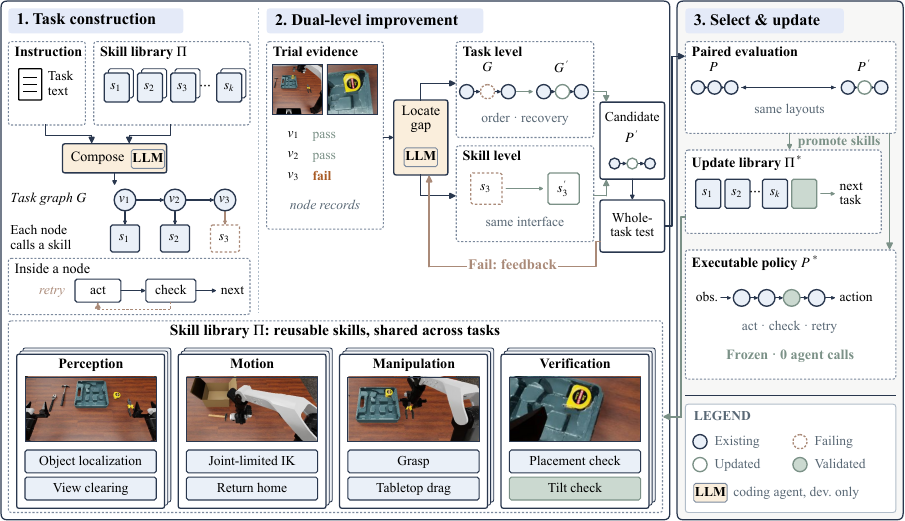}
\caption{ReSkill constructs and improves an executable task policy.
(1) A coding agent composes a task graph from the instruction and the skill library; each node calls a library skill and runs an act--check--retry loop.
(2) Node records from trial execution locate the failing node, prompting task-level graph revision or a skill-level implementation change behind the same interface; each candidate program is evaluated on the whole task, and failures return as feedback.
(3) Candidates that pass development and robustness evaluation are compared with the incumbent on the same test layouts; the retained program becomes a frozen executable policy with no run-time agent calls, and validated skills are promoted to the library for later tasks.
Bottom: the shared skill library groups reusable skills into perception, motion, manipulation, and verification, each shown with an original simulation example; the shaded entry marks a validated update.}
\label{fig:overview}
\end{figure}

\FloatBarrier
\subsection{Experimental and technical design}
\label{sec:m-setting}
\label{sec:method-setting}

\begingroup\par
We study how to construct a reusable task program for a new compositional manipulation task while concentrating development resources on the parts that existing capabilities do not cover.
Given a task description $\ell$ (a natural-language instruction and a description of the success criterion), the robot's observation and control interfaces, and optionally one demonstration $\tau^D$, the system outputs a task program
\begin{equation}
    \mathcal P=\bigl(G,\ \Pi,\ \Lambda\bigr),
    \label{eq:package}
\end{equation}
where $G$ is an object-centric task graph, $\Pi$ is a versioned skill library, and $\Lambda$ records node--skill bindings, parameters, and budgets.
The goal is for the same program to handle variation in initial object positions, object counts, and geometry within an object category while keeping the task goal and success criterion fixed.

At run time, the program may use only observations: head and wrist camera images, camera calibration, joint states, the robot model, and instruction text.
Object poses, simulator state, and evaluation results are not inputs; success is determined solely by the environment's native criterion, and the controller cannot declare success itself.
On RoboDojo, we enforce this through three layers of input auditing---static code checks, interception of run-time file access, and an environment boundary that exposes only the control interface---and deploy the program as a separate policy-server process that exchanges observations and actions with the simulator only through the official client.

ReSkill organizes program construction and improvement at two scales (Figure~\ref{fig:overview}).
The task level determines which stages compose a task, how stages hand off to one another, and how to recover from failure; the skill level determines whether the skill called at each stage is sufficiently reliable and improves the skill itself when it is not.
Both scales use the same execution records, and revisions occur between episodes: within an evaluation, the task graph and skills remain frozen, and only the closed-loop checks within nodes retry from new observations.
Revisions at both scales enter the next program version only after the multi-stage evaluation in Section~\ref{sec:method-revision}.
\par\endgroup

\FloatBarrier
\subsection{Object-centric task graph}
\label{sec:m-graph}
\label{sec:method-graph}

\begingroup\par
\textbf{Nodes and edges.}
The task graph is $G=(V,E)$, with nodes
\begin{equation}
    v_i=\bigl(s_i,\ \mathcal R_i,\ P_i,\ T_i,\ B_i\bigr),
\end{equation}
where $s_i$ is the skill or program called by the node, $\mathcal R_i=(\mathrm{Actor},\mathrm{Object},\mathrm{Goal/Reference})$ specifies the object roles, $P_i$ and $T_i$ are the entry and completion conditions, respectively, and $B_i$ is the action and retry budget.
Executing a node returns
\begin{equation}
    r_i=(\text{status},\ \text{cause},\ \text{evidence},\ \text{actions}),\qquad \text{status}\in\{\textsc{Pass},\textsc{Fail},\textsc{Unknown}\}.
\end{equation}
Edges $E$ specify a successor for each of the three outcomes: another node, termination, or return home.
A node returns \textsc{Unknown} when a required observation is missing, which does not count as success.
Execution ends upon reaching a terminal node, exhausting the action budget, exceeding a node's limit on consecutive unsuccessful attempts, or exceeding the total transition limit.
The status, cause, evidence, action count, and duration of each node visit are recorded as the common basis for failure attribution at both scales.

\textbf{Executor bindings.}
Nodes use two kinds of executors: stages with clear geometric goals---approach, transport, release, push, and return---are written directly as programs that calculate motion targets from current object relations, while interactions with reliable implementations in the library---such as grasping with empty-grasp verification and placement with offset verification---reuse skills.
Both kinds are implemented in code, with no new policy network trained for an individual task; the only end-to-end learned model used by a program is the segmentation model SAM~3, while all other perception, motion, and verification use geometric calculations and rules.

\textbf{Graph structure.}
Starting from fine-grained nodes with large-scale path search, nearly all task graphs converged during development to a few coarse nodes with an act--observe--retry loop inside each node.
A typical structure is survey $\rightarrow$ per-object work loop $\rightarrow$ return home; the work loop selects a target, grasps, verifies, transports, places, and checks again, with retries, arm switching, or hand-overs inside the node when an action fails.
Any failed or unknown node routes to the return-home node, ensuring that enough actions remain for both arms to return.
Tasks with strong structural dependencies retain more nodes: build tower, for example, has nine nodes organized by the support relations between parts.
The graph handles stage order, hand-off semantics, and recovery routes, while skill composition takes place within nodes; this division allows task-level revision and skill-level improvement to proceed separately.
\par\endgroup

\FloatBarrier
\subsection{Skill level: organizing, improving, and reusing skills}
\label{sec:m-skill}
\label{sec:method-skills}

\begin{figure}[!htbp]
\centering
\includegraphics[width=\linewidth]{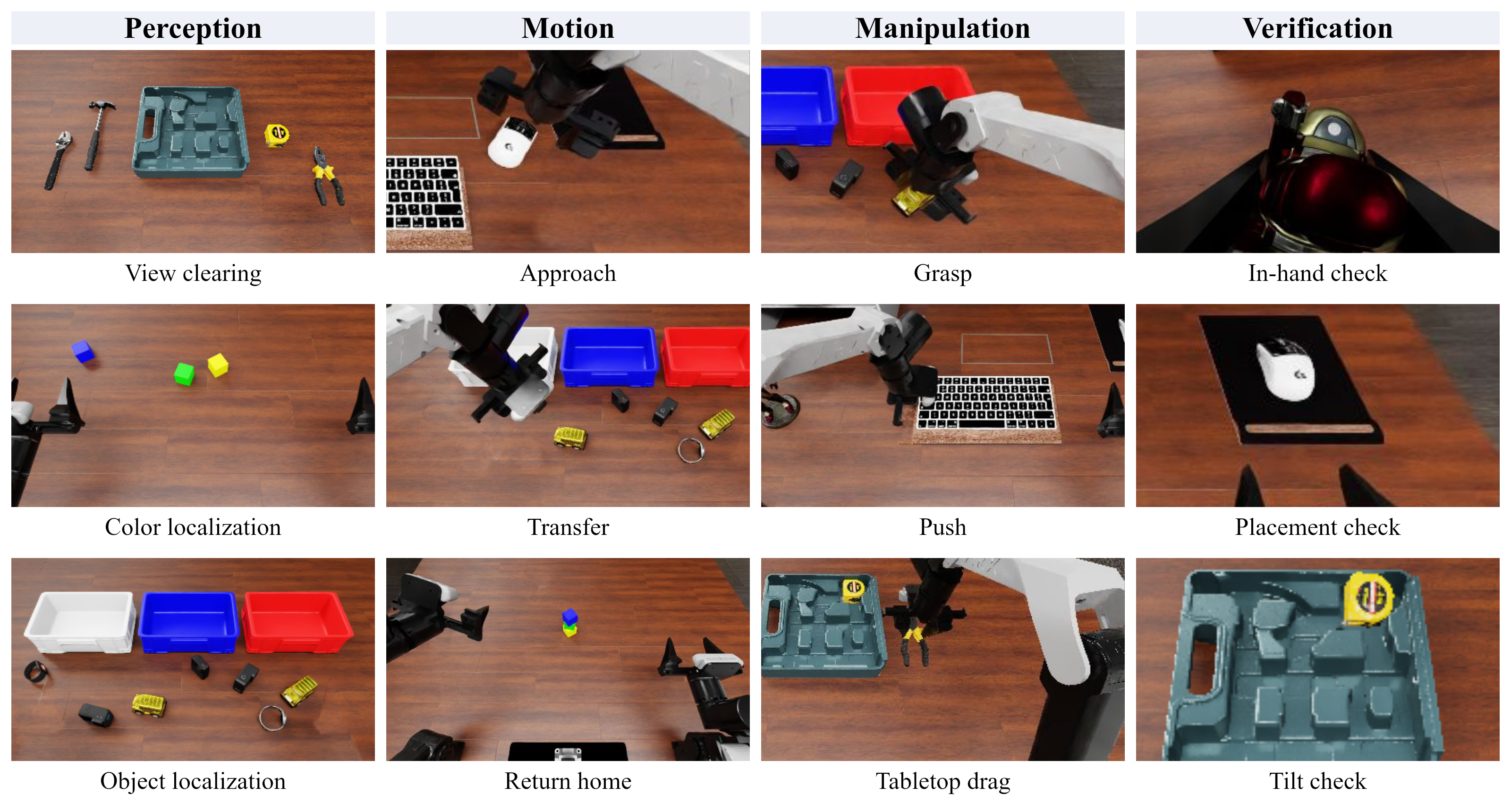}
\caption{Representative examples from the four skill categories.
Columns show perception, motion, manipulation, and verification.
The images show inputs, execution stages, or inspected states associated with the named skills; these associations are checked against execution traces and the corresponding frozen program implementations.}
\label{fig:skill-gallery}
\end{figure}

\begingroup\par
\textbf{Skill organization.}
The skill library $\Pi$ contains four categories: \emph{perception} skills convert images and depth into object geometry, including depth clustering using the table plane and height bands, point clusters within segmentation masks, color and shape discrimination, and moving the arms away before observing to avoid self-occlusion; \emph{motion} skills convert target poses into joint trajectories, including damped least-squares inverse kinematics seeded at the current joints and constrained to the real joint limits \cite{wampler1986dls}, Cartesian straight-line interpolation, and whole-arm collision checks; \emph{manipulation} skills carry out interactions, including vertical or tilted grasping, lift--translate--forward-tilt-and-lower placement, pushing, dragging, and bimanual hand-overs; and \emph{verification} skills judge interaction success through visual empty-grasp checks after lifting, offset and tilt checks after placement, and pre-grasp checks that the target remains in place.
Each skill records its role interface, admissible entry states, termination conditions, and version.
Figure~\ref{fig:skill-gallery} shows representative inputs, execution stages, and inspected states for these four categories.

\textbf{Skill improvement.}
Skill-level improvement is driven by failure evidence across tasks.
Execution records attribute each failure to a specific node and cause, such as an empty grasp, tilt after placement, or a missing observation; when the same skill fails for the same cause across tasks or layout types, the agent revises that skill individually without changing the task graph.
For example, objects swinging around the gripper during transport and landing off target led to a rule that grasps tall objects above their center of mass; objects appearing to be in their slots but not scoring, diagnosed from video as not lying flat, led to a verification skill that measures tilt after placement and presses the object flat or places it again; and the arms spreading the box walls and pushing the box away during insertion led to a release rule that opens the gripper only to the grasp width plus 1~cm inside the box, then opens it fully after lifting vertically above the rim.
An improved skill is first screened on the layouts where it failed and then undergoes full-task evaluation with the task program.

\textbf{Skill reuse.}
Skills are reused across tasks in three ways: generic implementations in the shared library that tasks call directly; cross-task transfer, which copies a validated skill implementation together with its parameters and verification logic into another task's program, either verbatim or with rewriting; and a cross-task lessons file, maintained by the orchestrator, that records rules confirmed by native evaluation together with their source tasks---for example, ``move both arms out of the head camera's view before observing'' and ``grasp tall objects above their center of mass''---and serves as a starting point for every new candidate.
New tasks start from the existing task with the closest skill requirements rather than an empty skeleton, retaining its task-graph structure and skill implementations and changing only what the new task requires.
Section~\ref{sec:exp-reuse} quantifies implementation reuse across task-specific modules.

\textbf{Skill replacement.}
The skill-level interface also allows an entire class of skills to be replaced while the task graph remains unchanged.
One example is motion: the generic inverse-kinematics solver used early in development did not use the current joints as its seed, often jumped between solution branches, and returned solutions outside the real joint limits that swung the arm backward, causing motion-planning failures or planning times of several hundred seconds.
We replaced it with damped least-squares inverse kinematics seeded at the current joints and constrained to the real joint limits \cite{wampler1986dls}, with planning and execution sharing the same solver; each task's graph and other skills remained unchanged, while planning time fell from several hundred seconds to 1--15~s.
Another example is placement: the original tool-storage skill carried tools through the air and released them above their slots, allowing slender tools to swing and miss the slots; it was replaced by a skill that slides the tool along the table to the outside of the box wall, lifts it over the wall at the lowest feasible height, and lowers it vertically, with graph nodes and hand-off conditions unchanged (Section~\ref{sec:exp-case-tools}).
\par\endgroup

\FloatBarrier
\subsection{Task level: failure-driven graph revision and selection}
\label{sec:m-task}
\label{sec:method-revision}

\begingroup\par
\textbf{Organizing failure evidence by stage.}
In round $r$, program $\mathcal P^{(r)}$ produces execution records $\mathcal D^{(r)}$ on development layouts, including node-level status and causes, multi-view visual traces sampled every ten actions, and the observations actually received by the controller.
The agent follows the graph to the earliest node that deviates and proposes a revision in light of its predecessor states:
\begin{equation}
    \mathcal P^{(r+1)}=\operatorname{Revise}\bigl(\mathcal P^{(r)},\ \mathcal D^{(r)}\bigr).
\end{equation}
Task-level revisions affect stage decomposition (splitting long stages or inserting re-observation or hand-over nodes), hand-off conditions (rewriting completion conditions to match the native criterion), skill bindings (switching a stage to another grasping or placement skill), and recovery routes (retrying, switching arms, or putting an object down and re-grasping after failure).
Skills themselves are revised at the skill level as described in Section~\ref{sec:method-skills}; the task level reorganizes execution through existing interfaces.

\textbf{Multi-stage evaluation.}
Each candidate is first screened on three development layouts; if it is no worse than the current best, it is evaluated on the full development set using the task score
\begin{equation}
J=100\times\mathrm{SR}_{\mathrm{DEV}}+\overline{S}_{\mathrm{DEV}},\qquad 0\le J\le 200,
\label{eq:J}
\end{equation}
where $\mathrm{SR}_{\mathrm{DEV}}$ is full-task success on the development set and $\overline{S}_{\mathrm{DEV}}$ is the mean native partial-completion score from RoboDojo (0--100, reflecting progress through the task).
We call $J$ the task score: the maximum of 200 means that every development episode succeeds, and partial-completion scores distinguish versions with the same success count.
Promising candidates are then evaluated on two groups of robustness layouts, after which the orchestrator compares each candidate with the incumbent on paired test layouts and decides whether to adopt it.
The test set is used for model selection, and its execution records are hidden from the development agent; once a task's test failures have been inspected, selection switches to a second test set.
Final validation requires evaluating the selected program on layouts that were not used in revision or model selection.

\textbf{Adoption rules.}
We progressively tightened the adoption rules to require improvements beyond the development layouts.
The development set contains only eight similar layouts, and candidates repeatedly improved on development layouts without improving on test layouts, or even regressed there; for such tasks, adoption therefore requires a strict improvement on the model-selection test set without regression on robustness layouts.
When the incumbent already succeeds on every test layout, the challenger must be strictly better on robustness layouts.
When two versions differ by at most four successes in a full evaluation of 64 episodes, each must be rerun at least twice on layouts where their outcomes disagree, and adoption uses the aggregate comparison rather than individual episodes.

\textbf{Parallel exploration.}
Several structurally different candidates may coexist for a task---for example, one that drags a box closer before placement and another that stages objects outside the box---and each iterates within its budget; candidates that fail to surpass the incumbent are stopped, but validated local changes can be transferred back to the main program.
\par\endgroup

\FloatBarrier
\subsection{Implementation}
\label{sec:m-impl}
\label{sec:method-impl}

\begingroup\par
The learning orchestrator uses Claude Code with Claude Opus 5.5 at two levels.
The orchestrating agent is a long-running session that checks the queue and candidate progress every 30 minutes, launches and stops candidates, runs candidate comparisons, decides adoption, maintains the cross-task lessons file, and promotes shared skills across tasks; development agents edit their own candidate programs and submit evaluations.
Humans provide only occasional directional instructions, such as stopping investment in demonstration-assisted candidates or adding tasks; they neither edit program code nor decide adoption.
Each candidate has an independent workspace containing its code and notes on hypotheses, changes, and results; when a session exhausts its context, a new session starts in the same workspace and continues from the notes and completed evaluations, making the notes the only memory across sessions.
All evaluations run on GPUs through a shared job queue, and each episode's native result, execution record, visual trace, and input-audit report are saved.
The only learned model used at run time is the segmentation model SAM~3 \cite{carion2025sam3}, with public weights kept frozen.
\par\endgroup

\FloatBarrier
\subsection{Experimental design and evaluation}
\label{sec:exp-setup}

\label{sec:m-design}

\textbf{Tasks and platforms.}
The experiments use two task groups.
Ten RoboDojo tasks \cite{chen2026robodojo} run on a bimanual ARX X5 robot in Isaac Sim, covering stacking, language-conditioned manipulation, classification, organization, and storage with different stage structures; each has a native action limit of 200--1300 control steps, with each step lasting 1/25~s, and a native success criterion.
Eight real-world tasks are evaluated in rebuilt MuJoCo environments and on the Xtrainer robot, with the same task goals and separately reported results.

\textbf{Observations and run-time configuration.}
RoboDojo controllers use only RGB-D images from the head and both wrist cameras, camera calibration, joint states, the robot model, and instruction text; they read neither simulator state nor measured finger opening, because the official observation interface provides only commanded finger values.
The coding agent participates only during development; the deployed program is frozen and runs as a separate policy-server process, exchanging observations and actions with the simulator only through the official client and making no external large-model calls at run time.
All main results are from \emph{final versions} with zero run-time API calls.

\textbf{Real-to-sim-to-real development.}
For each real-world task, we use one teleoperated demonstration and an agent-assisted real-to-sim process to reconstruct the corresponding robot, objects, and initial scene in MuJoCo, construct and revise the task graph in the rebuilt environment, and then deploy the frozen program on the real robot.
This process supplies a development environment for real-world tasks and is not itself part of ReSkill.
SAM~3 provides object masks in real images.

\label{sec:m-stats}

\textbf{Splits and protocol.}
RoboDojo evaluation uses development layouts to score and screen candidates, robustness layouts to assess changes across initial conditions, and a test set for orchestrator-led model selection.
The test-set execution records are hidden from development agents. We use \emph{test} to denote model-selection evaluations and reserve \emph{validation} for final reporting after selection.
Development uses the task score $J$ defined in Section~\ref{sec:method-revision}: it adds full-task success (0--100) to the native RoboDojo partial-completion score (0--100, reflecting progress through the task), giving a range of 0--200, with the maximum meaning that every development episode succeeds.
The RoboDojo results summarize the recorded evaluations of retained programs at seed 1, layouts 10--54. These records combine robustness and model-selection evaluations, including episodes used to select the retained versions; they therefore measure performance during development rather than independent final validation.
Full-task success uses the native criterion and counts episodes, including every repeat on the same layout in the denominator; the reported average is computed by first calculating the success rate of each task and then averaging these rates with equal task weights.

\textbf{Comparison methods.}
The RoboDojo Agent-as-Policy (Astra) and $\pi_{0.5}$ columns reproduce full-task success from the official evaluation report \cite{zhang2026astraeval}: Astra uses a zero-shot setting, \texttt{medium} reasoning effort, one seed, and 50 episodes per task; $\pi_{0.5}$ uses the leaderboard snapshot of 10 September 2026 included in the report.
These public results reflect each method's original inputs, training data, and evaluation protocol and are not fully comparable with our measurements.
We also report each task's initial task graph on the development set---the first version written from scratch by the development agent and evaluated on the full development set---to measure improvement during development.
For the real-world tasks, we report only ReSkill's success (Section~\ref{sec:exp-real}).

\FloatBarrier
\section{Results and Discussion}
\label{sec:results}
\label{sec:experiments}
\begingroup
\clubpenalty=10000
\widowpenalty=10000

The experiments address four questions: (1) how well the task programs built by ReSkill complete simulated and real-world tasks, and at what execution cost (Sections~\ref{sec:exp-system} and~\ref{sec:exp-real}); (2) how task- and skill-level improvements raise success, how much each revision changes, and how failure types evolve (Sections~\ref{sec:exp-improve} and~\ref{sec:exp-locality}); (3) how extensively skills are reused across tasks (Section~\ref{sec:exp-reuse}); and (4) what a single demonstration and development cost contribute (Sections~\ref{sec:exp-demo} and~\ref{sec:exp-cost}).
A tool-storage case study traces local skill revision and selection (Section~\ref{sec:exp-cases}).

\FloatBarrier
\subsection{Full-task execution performance}
\label{sec:exp-system}

\textbf{RoboDojo.} Figure~\ref{fig:robodojo-tasks} shows the ten task scenes, and Table~\ref{tab:main} summarizes their execution performance.
For each task, we report the adopted version with the highest success rate on its recorded evaluation layouts among versions that make no external model calls at run time.
Across the ten tasks, these retained programs succeed in 282 of 337 recorded evaluation episodes, with an average success rate of 83.6\% across tasks; the initial task graphs reach only 30.0\% on the development set.
Build tower and language-conditioned block stacking succeed in every reported episode (37/37 and 27/27); block stacking, bowl stacking, general pickup, object classification, and tool storage reach 87--97\%; language-conditioned classification reaches 74.1\%, box packing 63.0\%, and table organization 41.7\%.
The two hardest tasks fail in characteristic ways.
In box packing, failures concentrate on the last object, which touches the box wall or a previously placed object and ends up outside the orientation criterion.
In table organization, the figurine and the alarm clock topple or land outside their target regions, and most episodes exhaust the action limit.

\begin{figure}[!htbp]
\centering
\includegraphics[width=\linewidth]{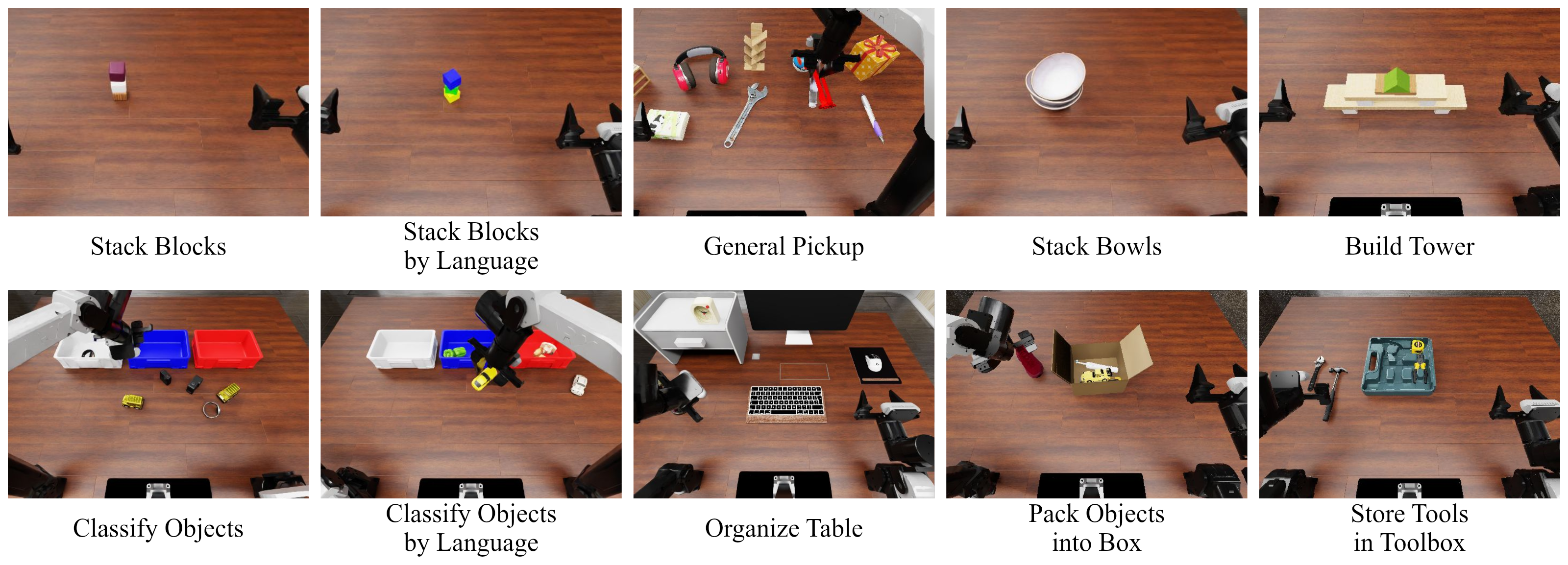}
\caption{Ten RoboDojo simulation tasks.
Representative head-camera views from recorded ReSkill executions.
The two rows follow the task order in Table~\ref{tab:main}.}
\label{fig:robodojo-tasks}
\end{figure}

For reference, the public report lists average success rates of 36.0\% for GPT-6 Astra used directly as the policy (Agent-as-Policy) and 9.6\% for the end-to-end $\pi_{0.5}$ on the same ten tasks.
Because their inputs, training data, and protocols differ from ours, these numbers indicate only the order of magnitude.
The difference in execution cost is more direct.
Agent-as-Policy queries a large model at every control decision.
A separate study on RoboLab \cite{yang2026robolab} reports mean Astra-call latencies of 11.50--18.56~s for its default policy configurations, with inference accounting for 94.9--97.8\% of estimated episode duration and estimated API costs of US\$1.96--36.40 per successful episode \cite{yang2026astranomy}.
ReSkill's final programs make no external model calls and incur no API cost at run time.
The task graphs and skills retained from development are available for reuse by subsequent tasks (Section~\ref{sec:exp-reuse}).

\begin{table}[!htbp]
\caption{Full-task success (\%) and execution cost on ten RoboDojo tasks.
Astra (Agent-as-Policy) and $\pi_{0.5}$ use success rates from the official report under different protocols and are shown for reference.
``Initial'' is development-set success for the first version the development agent evaluated in full.
``Succ.'' and ``SR'' report successes/episodes and success rate on the recorded robustness and model-selection evaluations (seed 1, layouts 10--54). All retained programs make zero run-time API calls.
``Steps'' gives the median control-step count over the native step limit.
The average is taken over the ten task success rates, with equal task weights.}
\label{tab:main}
\label{tab:exp-robodojo}
\centering
\footnotesize
\setlength{\tabcolsep}{4pt}
\begin{tabular*}{\linewidth}{@{\extracolsep{\fill}}lcccccc@{}}
\toprule
 & \multicolumn{2}{c}{Public report (\%)} & & \multicolumn{3}{c}{ReSkill retained program} \\
\cmidrule(lr){2-3}\cmidrule(lr){5-7}
Task & Astra & $\pi_{0.5}$ & Initial (\%) & Succ. & SR (\%) & Steps \\
\midrule
Stack blocks & 86.0 & 6.7 & 25.0 & 31/32 & \textbf{96.9} & 178/550 \\
Stack blocks (lang.) & 40.0 & 0.0 & 25.0 & 27/27 & \textbf{100.0} & 202/400 \\
General pickup & 84.0 & 12.0 & 100.0 & 51/57 & \textbf{89.5} & 57/200 \\
Stack bowls & 62.0 & 38.0 & 37.5 & 21/22 & \textbf{95.5} & 248/800 \\
Build tower & 2.0 & 24.0 & 37.5 & 37/37 & \textbf{100.0} & 722/1050 \\
Classify objects & 56.0 & 12.7 & 50.0 & 22/25 & \textbf{88.0} & 542/1100 \\
Classify (lang.) & 30.0 & 0.0 & 12.5 & 20/27 & \textbf{74.1} & 829/1100 \\
Organize table & 0.0 & 0.0 & 0.0 & 15/36 & \textbf{41.7} & 1000/1000 \\
Pack box & 0.0 & 2.7 & 12.5 & 17/27 & \textbf{63.0} & 973/1300 \\
Store tools & 0.0 & 0.0 & 0.0 & 41/47 & \textbf{87.2} & 566/900 \\
\midrule
Average (\%) & 36.0 & 9.6 & 30.0 & 282/337 & \textbf{83.6} & -- \\
\bottomrule
\end{tabular*}
\end{table}

\textbf{Real-world tasks in simulation.} On the eight real-world tasks rebuilt in simulation, each task is evaluated ten times, and ReSkill succeeds in 72 of 80 trials (90.0\%; first row of Table~\ref{tab:real}).
It succeeds in all ten trials of six tasks: apple to bowl, a single red cup to box, a single black spoon to plate, yellow cup into blue cup, drawer opening only, and a single yellow block to bowl.
Bimanual sorting of a single yellow block and sweeping with a single brush are the hardest, at 7/10 and 5/10: the former requires a block hand-over between the two arms, and the latter requires the brush to keep pushing a block into a dustpan over a long contact phase.

\FloatBarrier
\subsection{Real-robot execution}
\label{sec:exp-real}

We deployed the programs developed in the rebuilt scenes on the Xtrainer robot and evaluated the same eight tasks (second row of Table~\ref{tab:real}).
Task graphs, programs, and skills were frozen before testing, remained unchanged during execution, and no external large model was called at run time.

\textbf{Test protocol and success criteria.} Each task used a fixed set of object configurations and initial states, and the scene was reset before every trial.
A trial ran continuously from the initial state to the goal; a mid-trial reset, human takeover, or exceeding the action budget or time limit counted as failure.
Success required the intended final state: picked or sorted objects inside the designated container or region, the cups nested, the drawer opened to the required extent, and the swept block inside the dustpan.
Each task was tested ten times, with at most 2400 control steps per trial; exceeding 120~s counted as a timeout. All criteria were fixed before testing.

\begin{table}[!htbp]
\caption{ReSkill successes on eight real-world tasks in rebuilt simulation and on the real robot (ten trials per task).}
\label{tab:real}
\label{tab:exp-real}
\centering
\small
\setlength{\tabcolsep}{3pt}
\begin{tabular*}{\linewidth}{@{\extracolsep{\fill}}lccccccccc@{}}
\toprule
 & Apple & Red cup & Black spoon & Cup & Drawer & Block & Bimanual & Sweep & Total \\
 & to bowl & to box & to plate & nesting & & to bowl & sorting & & \\
\midrule
Rebuilt simulation & 10 & 10 & 10 & 10 & 10 & 10 & 7 & 5 & 72/80 \\
Real robot & 9 & 9 & 8 & 8 & 9 & 9 & 6 & 4 & 62/80 \\
\bottomrule
\end{tabular*}
\end{table}

On the real robot, ReSkill succeeded in 62 of 80 trials (77.5\%), 12.5 percentage points below the rebuilt simulation, losing one or two trials per task.
Classifying the 18 failures by the earliest failing step, six came from perception and segmentation, where reflections or mutual occlusion caused missing SAM~3 masks or confused object roles so that later skills started from a wrong entry state; eight occurred during local interaction, mainly the brush losing contact with the block while sweeping and the block slipping during the bimanual hand-over; and four were hand-off errors in which a completion condition was met too early or too late under real sensing errors.
Perception and hand-off failures together account for ten, echoing the self-check/benchmark disagreement in RoboDojo: improving real-world execution also requires better state judgment beyond local skills.

Figure~\ref{fig:exp-real-sequences}(a) pairs the source demonstration scenes with their MuJoCo reconstructions; panel (b) shows how grasping, lifting, transfer, placement, and retreat appear across four real-world sequences.
Block collection repeats a pick-and-place loop, drawer storage adds a pulling stage, and the bimanual sequence transfers the block between arms through an intermediate placement on the table.

\begin{figure}[!htbp]
\centering
\includegraphics[width=\linewidth,height=0.82\textheight,keepaspectratio]{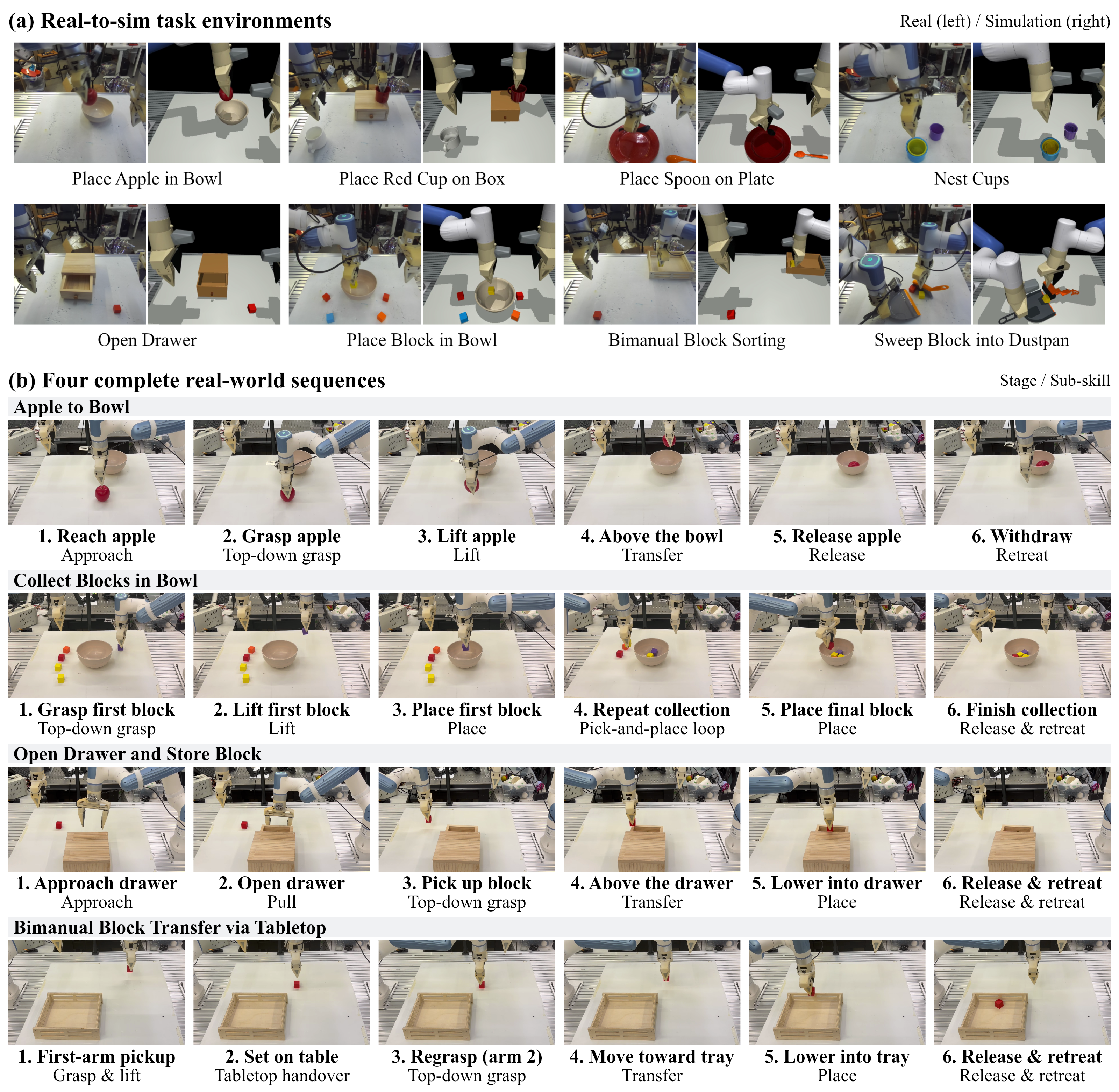}
\caption{Real-world task environments and four execution sequences.
(a) Eight task families, each pairing a real demonstration frame (left) with its MuJoCo reconstruction (right).
(b) Selected keyframes spanning apple-to-bowl, block collection, drawer storage, and bimanual tabletop transfer, using one video per task.
The upper label beneath each frame names the stage; the lower label identifies the visible sub-skill.
These qualitative sequences include multi-object operations; quantitative task variants are defined in Table~\ref{tab:exp-real}.
The sweeping reconstruction in (a) uses a single yellow block.}
\label{fig:real}
\label{fig:exp-real-sequences}
\label{fig:xtrainer-real2sim}
\end{figure}

\FloatBarrier
\subsection{Skill improvement: success and failure types over successive revisions}
\label{sec:exp-improve}

\textbf{Improvement through local revision.}
Across the ten tasks, average success on the fixed development layouts rises from 30.0\% for the initial task graphs to 86.2\% for the final adopted programs.
Block stacking, language-conditioned block stacking, bowl stacking, build tower, and object classification reach or approach full success within 3--7 evaluated versions, whereas table organization, box packing, tool storage, and language-conditioned classification require 25--69 versions.
The harder tasks improve through a sequence of local changes.
In tool storage, the agent first revises the placement path to clear the box wall, then adjusts grasp-point selection for the wrench and pliers, and finally adds a post-placement tilt check followed by pressing or re-grasping.
Each candidate is evaluated as a complete task program before adoption, allowing successful changes to accumulate while preserving the stages that already work.

Local execution success does not imply task completion.
Figure~\ref{fig:failures} shows how failure types change during development.
For example, tool storage succeeds in 176 of 202 late episodes, yet 24 of the 26 remaining failures pass the node checks, showing that the current local completion criteria still miss errors that prevent full-task success.
This finding indicates that self-improvement must strengthen both skill execution and the checks for node completion and stage hand-offs, so that subsequent actions start from the correct intermediate states.
Full-task evaluation therefore provides essential feedback for local revision: it exposes errors missed by node checks and guides further changes to skills, completion conditions, or recovery logic.

\begin{figure}[!htbp]
\centering
\includegraphics[width=\linewidth]{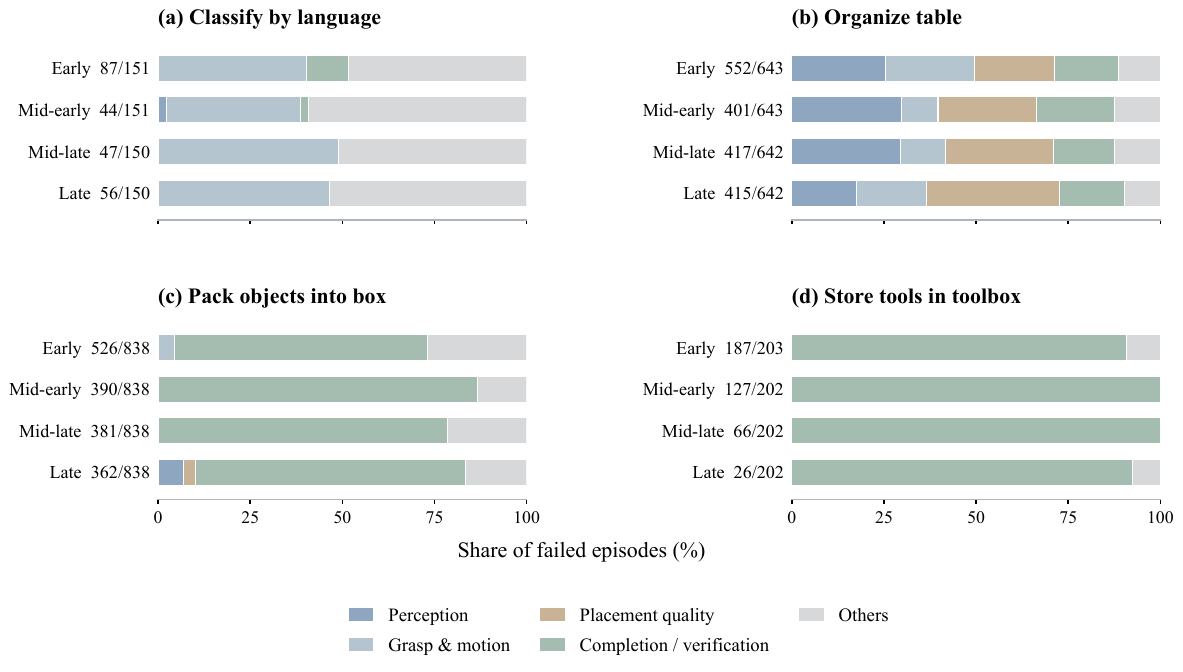}
\caption{Composition of failed episodes during development.
All evaluated candidates on development and robustness layouts are divided into four consecutive, equally sized groups by episode completion order.
Bars normalize the five cause groups over failed episodes only; labels give failures/total episodes, so changes in the number of failures remain visible.
These periods describe the exploration process, rather than successive retained policies.}
\label{fig:failures}
\end{figure}

\FloatBarrier
\subsection{Revision locality}
\label{sec:exp-locality}

To test directly whether improvement is confined to what is insufficient, we analyzed every code snapshot of the ten tasks.
Two consecutive snapshots of the same candidate program form one revision, giving 774 revisions.
For each revision we parsed all functions of both versions (excluding the evaluation harness), compared their abstract syntax trees, and counted added, removed, and modified functions.
Functions that build the task graph and the node entry functions they reference were assigned to the \emph{task level}, all other functions (perception, motion, manipulation, and verification skills) to the \emph{skill level}; revisions that changed only parameter files were counted separately.
This assignment follows code structure and is approximate, because node entry functions often contain skill-composition logic.
For the 317 revisions whose two versions each have at least eight verified episodes on the same layouts, we also compared full-task success on those layouts.

Revisions are highly local.
The median program comprises 578 functions and about 11{,}700 lines of code, whereas the median revision changes 3 functions---0.5\% of the functions and 0.3\% of the lines---and 90\% of revisions change at most 1.9\% of either.
Table~\ref{tab:locality} breaks revisions down by type: skill-only and two-level revisions each account for about 39\%, parameter-only revisions for 16\%, and task-only revisions for 6\%.
Among revisions that raised success, two-level revisions contributed 48\% of the total gain and skill-only revisions 36\%.
This agrees with the graph shape described in Section~\ref{sec:m-graph}: once a task graph has converged to a few coarse nodes, most improvement happens in the skills inside nodes, and task-level changes usually come together with matching skill changes.

\begin{table}[!htbp]
\caption{Revision statistics by change type.
``Funcs.'' is the median number of changed functions; ``Compared'' counts revisions whose two versions each have at least eight verified episodes on the same layouts; ``Improved'' counts revisions with higher success; ``$\Delta$SR'' is the mean change in full-task success across paired episodes for those compared revisions, in percentage points; and ``Gain share'' is each type's share of the total positive success-rate improvement.}
\label{tab:locality}
\label{tab:exp-locality}
\centering
\small
\begin{tabular*}{\linewidth}{@{\extracolsep{\fill}}lrrrrrr@{}}
\toprule
Change type & Revisions & Funcs. & Compared & Improved & $\Delta$SR (pp) & Gain share \\
\midrule
Skill level & 303 & 2 & 119 & 58 & +2.8 & 36\% \\
Task level & 47 & 1 & 22 & 11 & +3.3 & 4\% \\
Both levels & 299 & 5 & 113 & 56 & +5.6 & 48\% \\
Parameters only & 125 & 0 & 63 & 25 & +1.2 & 12\% \\
\midrule
Total & 774 & 3 & 317 & 150 & +3.5 & 100\% \\
\bottomrule
\end{tabular*}
\end{table}

Revisions also improve success on average.
Across the 317 comparable revisions, full-task success changed by $+3.5$ percentage points on average, and 150 revisions (47\%) raised it.
Two-level revisions brought the largest mean gain ($+5.6$ points) and parameter-only revisions the smallest ($+1.2$ points).
Individual revisions are small and not always effective; improvement accumulates over development through the multi-stage evaluation in Section~\ref{sec:method-revision}: versions are retained when paired comparisons satisfy the adoption rules, including improved robustness when test success is already saturated.

\FloatBarrier
\subsection{Skill reuse}
\label{sec:exp-reuse}

\textbf{Function-level reuse.} To measure reuse, we parsed function definitions in the task-specific modules of each adopted program and compared abstract syntax trees across tasks.
A function identical to a same-named function in another task counts as reused verbatim; a same-named function with a different body counts as reused with rewriting.
Besides the ten benchmark tasks, we include four RoboDojo tasks added midway through development (bottles to bin, ring-stacking toy, eggs to tray, and laptop-and-headphone storage), which are used to examine reuse in new tasks and are excluded from the ten-task benchmark and revision statistics.
Figure~\ref{fig:reuse} shows the result.
Of the 611 task-specific functions in the ten benchmark programs, 19\% are identical to a function in another task; of the 345 task-specific functions in the four later tasks, 39\% are identical, ranging from 33\% to 47\% per task.
Reuse follows clear lineages.
The lean closed loop of the object-classification program---pose representation, forward kinematics, joint-limited inverse kinematics, joint unwrapping, and grasp planning---was carried over verbatim to box packing and to the bottle, egg-holder, and laptop tasks, while the motion layer of language-conditioned block stacking---nearest-solution search, reachability checks, top-down grasp poses, distance-based arm selection, hand-over points, and return---was transplanted as a whole to the stacking toy.
Block stacking, bowl stacking, build tower, general pickup, and table organization were written early from different starting points, and reuse among them is mostly at the level of same-named, rewritten functions.

\begin{figure}[!htbp]
\centering
\includegraphics[width=\linewidth]{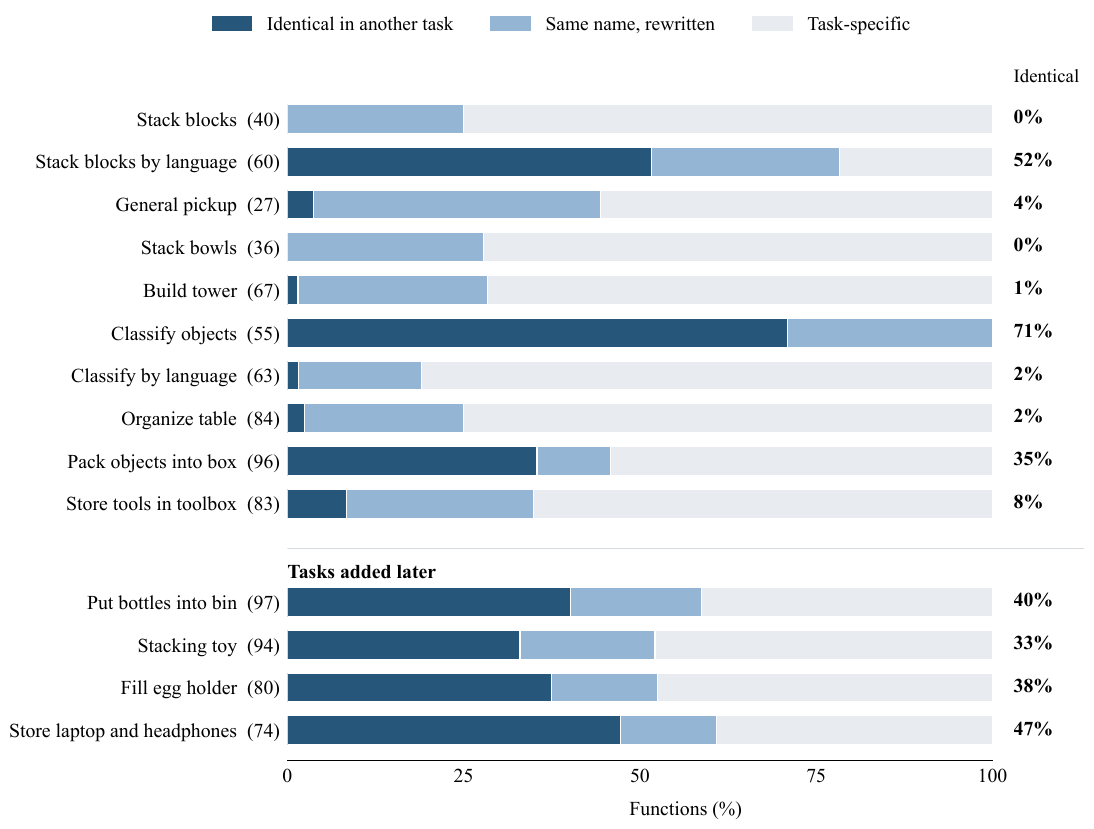}
\caption{Function-level skill reuse.
Each row covers function definitions in the task-specific modules of one adopted program (count in parentheses), split into functions identical to another task, same-named functions with rewritten implementations, and functions unique to the task; the number at right is the verbatim share.
The bottom four tasks were added during development to examine reuse in new tasks.}
\label{fig:reuse}
\end{figure}

\FloatBarrier
\subsection{The role of a demonstration}
\label{sec:exp-demo}

RoboDojo provides public demonstrations for several tasks.
To examine what one demonstration contributes, at the start of development we created a \emph{demonstration-assisted} candidate for each of build tower, object classification, block stacking, and bowl stacking.
The agent read the demonstration offline during development to extract stage order, placement points, release heights, and object order; the run-time program never reads it.
Two candidates that did not see the demonstration started at the same time, one with bounded path search and one with a lean act--observe--retry loop.
Table~\ref{tab:demo} compares them.

\begin{table}[!htbp]
\caption{Demonstration-assisted candidates compared with candidates started simultaneously without a demonstration, taking the faster or better of the two demonstration-free candidates.
Times are in hours from candidate creation: ``First success'' is the first successful full episode, and ``First DEV 8/8'' is the completion time of the first evaluation that succeeds on all eight development layouts; ``Best DEV'' includes only versions evaluated on the full development set; ``Best TEST'' gives the highest recorded successes/episodes on model-selection layouts.}
\label{tab:demo}
\label{tab:exp-demo}
\centering
\footnotesize
\begin{tabular*}{\linewidth}{@{\extracolsep{\fill}}lrrrrrrrr@{}}
\toprule
 & \multicolumn{2}{c}{First success (h)} & \multicolumn{2}{c}{First DEV 8/8 (h)} & \multicolumn{2}{c}{Best DEV} & \multicolumn{2}{c}{Best TEST} \\
\cmidrule(lr){2-3}\cmidrule(lr){4-5}\cmidrule(lr){6-7}\cmidrule(lr){8-9}
Task & Demo & None & Demo & None & Demo & None & Demo & None \\
\midrule
Build tower & 0.7 & 3.6 & 4.5 & 4.2 & 8/8 & 8/8 & 7/8 & 6/8 \\
Classify objects & 1.5 & 3.6 & 4.7 & 6.5 & 8/8 & 8/8 & 8/8 & 7/8 \\
Stack blocks & 0.7 & 0.5 & 2.1 & 3.2 & 8/8 & 8/8 & 7/8 & 8/8 \\
Stack bowls & 2.9 & 0.5 & 4.0 & 3.6 & 8/8 & 8/8 & 7/8 & 8/8 \\
\bottomrule
\end{tabular*}
\end{table}

A demonstration shortens the time to the first success in multi-stage tasks. For multi-stage tasks such as build tower and object classification, the demonstration directly supplies the stage order (pillars before the long board) and the placement style (release while tilting forward along the arm's radial direction).
The demonstration-assisted candidates first succeeded after 0.7 and 1.5~h, whereas the demonstration-free candidates needed 3.6~h in both tasks, about 5 and 2.3 times as long, respectively.
For simple stacking tasks, the demonstration brought no speed-up (0.7 versus 0.5~h for block stacking and 2.9 versus 0.5~h for bowl stacking).

The best recorded model-selection results are similar for these four task pairs: demonstration-assisted versus demonstration-free candidates achieve 7/8 versus 6/8 for build tower, 8/8 versus 7/8 for object classification, and 7/8 versus 8/8 for both block stacking and bowl stacking. Each pair differs by one successful layout.
The first all-success development version took 4.5 versus 4.2~h for build tower and 2.1 versus 3.2~h for block stacking.
We therefore stopped investing in demonstration-assisted candidates early in development, and their transferable geometric insights, such as the radial forward-tilt release, were transplanted as skills into other candidates.
A demonstration thus helps the agent write a task graph that succeeds once sooner, whereas the path from occasional to reliable success still runs through failure-driven revision.

\FloatBarrier
\subsection{Development cost}
\label{sec:exp-cost}

The development cost considered here is the coding agent's model usage while constructing, revising, evaluating, and selecting task programs.
The deployed policy is frozen and makes no run-time calls to the coding agent.

\textbf{Recorded model usage.}
The recoverable logs contain activity on five dates during the final development period, totaling 281 sessions, 24{,}699 model calls, and 16.26 million output tokens.
These totals describe the logged period rather than the complete development history.
The resulting code and skills remain available for reuse after this development expenditure.

\FloatBarrier
\subsection{Case study: local skill improvement in tool storage}
\label{sec:exp-cases}
\label{sec:exp-case-tools}

Tool storage shows how successive revisions to local skills improve whole-task success.
The task requires placing a tape measure, hammer, wrench, and pliers into their matching slots.
For elongated tools, the existing placement skill slides the tool along the table to the outside of the box wall, carries it over the wall at low height, and lowers it vertically.
Figure~\ref{fig:case-study} traces eight candidate versions from rounds 6--15 of one development branch, each evaluated on the same branch-specific set of 40 development layouts.
Throughout these rounds, the survey--store--home task graph retains its nodes, conditions, and edges, while revisions target perception, grasping, placement, and post-placement verification.

\begin{figure}[!htbp]
\centering
\includegraphics[width=\linewidth,height=0.78\textheight,keepaspectratio]{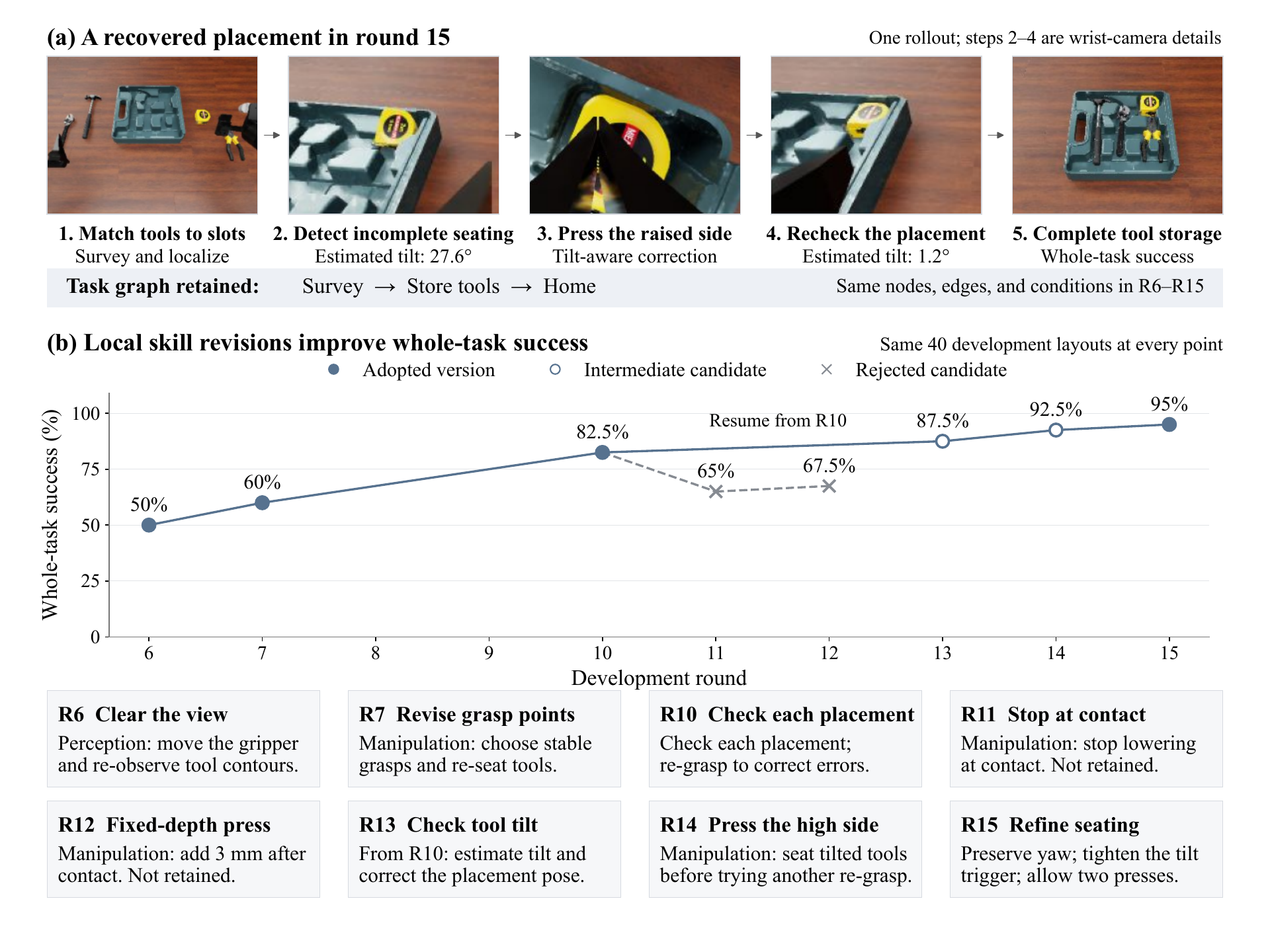}
\caption{Local skill improvement in tool storage.
(a) Frames from one round-15 rollout show localization, detection of a tilted placement, high-side pressing, rechecking, and completed storage. Frames 2--4 are wrist-camera details; tilt angles are computed from adjacent observations.
(b) Development results with simulator-measured depth, showing only candidates evaluated on the same 40 layouts. The horizontal axis retains the original round numbers; the vertical axis reports whole-task success. Lines indicate revision ancestry: round 13 resumes from round 10. Filled circles, open circles, and crosses denote adopted versions, intermediate candidates, and rejected candidates, respectively. The annotations list each round's skill changes; the macro task graph is unchanged throughout this window.}
\label{fig:case-study}
\label{fig:exp-case-study}
\end{figure}

\textbf{From pick-and-place to post-placement verification.}
By round 6, the program could move the gripper away and re-observe when a tool contour was truncated, achieving $20/40$ successes ($50.0\%$).
Round 7 used execution records to revise grasp locations and add re-seating, raising success to $24/40$ ($60.0\%$).
Attention then shifted to the state of each tool after release: round 10 checked position, orientation, and rim clearance after every placement and used the observed placement error to re-grasp the tool at its current location and return it to the intended pose.
Success reached $33/40$ ($82.5\%$), with post-placement observations now determining the next corrective action.

\textbf{Selecting revisions by whole-task outcomes.}
Rounds 11 and 12 revised plier placement by stopping descent on contact and then adding a fixed pressing distance, but success fell to $26/40$ ($65.0\%$) and $27/40$ ($67.5\%$), respectively, below round 10.
Both candidates were rejected, and development resumed from round 10.
This rejected branch shows why a change proposed to address a local symptom must still be evaluated in the complete task before it is retained.

\textbf{From repeated re-grasping to tilt-aware seating.}
The remaining failures exposed a more specific capability gap: a tape measure could enter its slot while leaning against the rim, and re-grasping at the same pose could preserve the tilt.
Round 13 added tilt estimation and triggered correction; round 14 pressed the raised side according to the tilt direction; and round 15 revised transport and gripping to preserve tool yaw, tightened the tilt trigger, and allowed a second press.
These candidates achieved $35/40$ ($87.5\%$), $37/40$ ($92.5\%$), and $38/40$ ($95.0\%$), respectively.
The rollout frames show this feedback loop in action: after detecting incomplete seating, the policy presses the raised side, rechecks an estimated tilt reduction from $27.6^{\circ}$ to $1.2^{\circ}$, and goes on to store all four tools.
The adopted round-15 version retains tilt verification and directed pressing as routines that subsequent executions can call directly.

\FloatBarrier
\endgroup

\FloatBarrier
\subsection{Discussion}
\label{sec:discussion}

\textbf{When selective improvement works.}
ReSkill focuses improvement on parts that repeatedly fail during execution: the median revision changes only 3 functions, and once a task graph converges to a few coarse nodes, most gains come from revising the skills within those nodes.
This works best under two conditions.
First, failures must be attributable to specific nodes and causes: node-level records of status, causes, and evidence allow the agent to decide whether to revise the task graph or a skill.
In tool storage, diagnosing the episodes in which self-checks passed but the native criterion failed as tools not lying flat led to a concrete verification skill and a substantial rise in success.
Second, the perception on which skills depend must be relatively stable: the placement, verification, and re-grasping skills accumulated during development all rely on dependable estimates of object geometry.
Object-centric semantic fields offer a complementary way to stabilize part-aware representations across observations and object instances \cite{sun2026semanticfields}; integrating these representations into perception skills is a direction for future work.

\textbf{Limits of selective improvement.}
Local revision may be insufficient when these conditions do not hold.
First, performance remains sensitive to the quality of the geometry estimates: the tool-storage placement skill reaches 41/47 with accurate depth geometry, whereas another lineage of the same task reaches 2/32 with less accurate depth geometry.
Second, reuse cannot supply a capability entirely absent from the library: laptop storage requires inserting a laptop at a specific orientation into a slot only about 3~mm wider than the laptop, and development went a long time without a full success because there was no reusable ancestor skill.
Third, in multi-object, long-sequence tasks such as table organization and box packing, remaining failures are distributed across placement quality, exhausted budgets, and criterion mismatch; each revision fixes only a small part, and progress slows over successive rounds.
High-precision contact and long sequences are the conditions hardest to cover with rule-based closed loops and most in need of local learning or online adaptation; enabling the agent to decide when to switch from editing code to learning a small policy is a natural next step for selective self-improvement.

\textbf{From simulation to self-improvement on real robots.}
Self-improvement in this work takes place mainly in simulation: RoboDojo development ran more than ten thousand episodes, and task graphs and skills for the real-world tasks were revised in rebuilt simulations before deployment on the real robot.
Simulation offers conditions that are difficult to obtain on hardware, including unrestricted resets, parallel execution, repeated evaluation of the same layout, and automatic scoring consistent with the task criterion.
Performing the same self-improvement directly on a real robot would require addressing automatic resets, success detection, and evaluation cost, and remains to be explored.
Our results nevertheless provide a reference: with sufficiently extensive evaluation, an agent starting from a task description and improving only what is insufficient can reach 83.6\% average success across ten tasks in recorded evaluations while leaving skills that transfer across tasks.
The next step is to determine which parts of this development loop can be transferred to real robots.

\FloatBarrier
\section{Conclusion}
\label{sec:conclusion}

We introduced ReSkill to identify capability gaps in compositional manipulation, improve the relevant parts, and retain skills for continued use.
An object-centric task graph organizes existing capabilities, while node-level execution records guide revision at two scales: the task level changes stages, hand-off conditions, and recovery routes, and the skill level improves perception, motion, manipulation, and verification routines.
Candidate revisions are retained only after screening, full-task evaluation, and paired comparisons on the model-selection test set.
The resulting policy executes independently, without an agent at run time.

Experiments show that local revisions can accumulate into improved full-task performance: retained ReSkill programs average 83.6\% success across ten RoboDojo tasks in recorded evaluations and succeed in 72 of 80 rebuilt-simulation trials and 62 of 80 real-robot trials across eight real-world tasks.
Across 774 revisions, the median change involves only three functions, and 39\% of the functions in later-added tasks directly reuse existing implementations.
Development thus leaves validated skills and execution logic that subsequent programs can build on.
The current improvement process relies mainly on simulation with repeatable evaluation and on programmatic skills supported by visual observations.
Future work will introduce local policy learning where these skills are insufficient and extend failure diagnosis, candidate validation, and skill retention to real robots.

\FloatBarrier
\section*{Acknowledgments}
\subsection*{Author Contributions}
Zheng Sun conceived the study, performed the experiments, and wrote the manuscript.
Zhihao Li, Zhuo Li, Xin Wang and Fei Chen contributed to the conception of the study and manuscript writing.
Wenrui Liu contributed to the experiments and manuscript writing.

\subsection*{Funding}
The authors received no funding for this work.

\subsection*{Competing Interests}
The authors declare that they have no competing interests.

\subsection*{Use of AI-assisted tools}
AI-assisted tools were used for language editing and polishing of the manuscript.
AI models also form components of the experimental system: a coding agent constructs and revises task programs during development, and a frozen segmentation model supplies object masks, as described in Methods.
The authors reviewed the AI-assisted text and take responsibility for the manuscript's content.

\subsection*{Data Availability}
Episode-level results, revision snapshots, and the scripts that produce every figure and table in this article will be made publicly available upon publication.

\begingroup
\setlength{\emergencystretch}{2em}
\printbibliography
\endgroup

\end{document}